\RequirePackage{fix-cm}
\documentclass[11pt,a4paper]{article}   
\usepackage[margin=1.0in]{geometry}
\usepackage[doublespacing]{setspace}                 
\usepackage{graphicx}
\usepackage{url}
\usepackage{apacite}

\begin{document}



\title{Predicting University Students' Academic Success and Major using Random Forests}



\author{C\'edric Beaulac         \and
        Jeffrey S. Rosenthal  
}




\date{\today}

\maketitle

\begin{abstract}
In this article, a large data set containing every course taken by every undergraduate student in a major university in Canada over 10 years is analysed. Modern machine learning algorithms can use large data sets to build useful tools for the data provider, in this case, the university. In this article, two classifiers are constructed using random forests. To begin, the first two semesters of courses completed by a student are used to predict if they will obtain an undergraduate degree. Secondly, for the students that completed a program, their major is predicted using once again the first few courses they have registered to. A classification tree is an intuitive and powerful classifier and building a random forest of trees improves this classifier. Random forests also allow for reliable variable importance measurements. These measures explain what variables are useful to the classifiers and can be used to better understand what is statistically related to the students' situation. The results are two accurate classifiers and a variable importance analysis that provides useful information to university administrations.

\bigskip

\noindent

\end{abstract}

\pagebreak

\section{Introduction}
\label{intro}

Being able to predict if a student is at risk of not completing its program is valuable for universities that would like to intervene and help those students move forward. Predicting the major that will be completed by students is also important in order to understand as soon as possible which program attracts more students and allocate resources accordingly. Since gathering data can be an expensive procedure, it would be useful being able to predict both of these things using data the university already possesses such as student records. Understanding which variables are useful in both of these predictions is important as it might help understand what drives student in taking specific classes.

\bigskip

Formally, these two prediction problems are classification ones. To solve these, a popular machine learning algorithm is used, a classification tree.  A classification tree is an easy to interpret classification procedure that naturally allows interactions of high degree across predictors. The classification tree uses the first few courses attempted and grades obtained by students in order to classify them. To improve this classifier, multiple trees are grown and the result is a random forest. A random forest can also be used to assess variable importance in a reliable manner. 

\bigskip

The University of Toronto provided a large data set containing individual-level student grades for all undergraduate students enrolled at the Faculty of Arts and Science at the University of Toronto - St. George campus between 2000 and 2010.  The data set contains over 1 600 000 grades and over 65 000 students. This data set was studied by Bailey et al. \citeyear{Bailey16} and was used to build an adjusted GPA that considers course difficulty levels. Here, random forest classifiers are built upon this data set and these classifiers are later tested.   

\bigskip

The contribution in this article is two-fold. First, classifiers are built and the prediction accuracy of those classifiers exceeds the accuracy of the linear classifiers thus making them useful for universities that would like to predict where their resources need to be allocated. Second, the variable importance analysis contains a lot of interesting information. Among many things, the high importance of grades in low-grading departments was noted and might be a symptom of grade inflation.
\section{Literature review}
\label{sec:2}

\subsection{Predicting success}

In this article a statistical learning model is established to predict if a student succeeds at completing an undergraduate program and to predict what major was completed. This statistical analysis of a higher education data set shares similarities with recent articles by Chen and Desjardins \citeyear{Chen08,Chen10} and Leeds and DesJardins \citeyear{Leeds15} as a new statistical approach will be introduced, a data set will be presented and policy making implications will be discussed. The task of predicting student academic success has already been undertaken by many researchers.  Recently Kappe and van des Flier \citeyear{Kappe12} tried to predict academic success using personality traits. In the meanwhile, Glaesser and Cooper \citeyear{Glaesser12} were interested in the role of parents' education, gender and other socio-economic metrics in predicting high school success. 

\bigskip

While the articles mentioned above use socio-economic status and personality traits to predict academic success, many researchers are looking at academic-related metrics to predict graduation rates. Johnson and Stage \citeyear{Johnson18} use High-Impact Practices, such as undergraduate research, freshman seminars, internships and collaborative assignments to predict academic success. Using regression models, they noted that freshman seminars and internships were significant predictors. Niessen and al. \citeyear{Niessen16} discuss the significance of trial-studying test in predicting student dropouts. This test was designed to simulate a representative first-year course and student would take it before admission. The authors noted that this test was consistently the best academic achievement predictor.

\bigskip

More recently, Aulck and al. \citeyear{Aulck16} used various machine learning methods to analyse a rather large data set containing both socio-economic and academic metrics to predict dropouts. They noted similar performances for the three methods compared; logistic regression, k-nearest neighbours and random forests.  The proposed analysis differs from the above-mentioned as it takes on the challenge to predict academic success and major using strictly academic information available in student records. The benefits of having classifiers built upon data they already own is huge for university administrations. It means university would not need to force students to take entry tests or relies on outside firms in order to predict success rate and major which is useful in order to prevent dropout or to allocate resources among departments. As noted by Aulck and al. \citeyear{Aulck16} machine learning analysis of academic data has potential and the uses of random forest in the following article aims at exploiting this potential.

\subsection{Identifying important predictors}
\label{gi}

Identifying and interpreting the variables that are useful to those predictions are important problems as well. It can provide university administrator with interesting information. The precise effect of grades on a student motivation lead to many debates and publications over the years (more recently \cite{Mills00,Ost10}). Because grades should be indicators of a student's abilities, evaluating the predictive power of grades in various departments is important. University administrators might want to know if grades in a department are better predictors than grades in other departments. Continuing on the point, it is also important to understand what makes the evaluations in a department a better indicator of students' success. Random forest mechanisms lead to variable importance assessment techniques that will be useful to understand the predictive power of grades variables. 

\bigskip

Understanding the importance ranking of grades in various departments can also enlighten us regarding the phenomenon of \textit{grade inflation}. This problem and some of its effect has been already discussed in many papers (\cite{Sabot91,Johnson03,Bar09} ) and it is consensual that this inflation differs from one department to another. According to Sabot and Wakeman-Linn, \citeyear{Sabot91} this is problematic since grades serve as incentives for course choices for students and now those incentives are distorted by the grade inflation. As a consequence of the different growths in grades, they noted that in many universities there exist a chasm in grading policies creating high-grading departments and low-grading departments. Economics, Chemistry and Mathematics are examples of low-grading departments while English, Philosophy and Political Science are considered high-grading. 

\bigskip

As Johnson mentions \cite{Johnson03}, students are aware of these differences in grading, openly discuss them and this may affect the courses they select. This inconsistency in course difficulty is also considered by Bailey and al. \citeyear{Bailey16} as they built an adjusted GPA that considers course difficulty levels. The accuracy of that adjusted GPA in predicting uniform test result is a great demonstration that courses do vary in difficulty. If some departments suffer
from grade inflation, the grades assigned in that department should be less tied to the actual student ability and therefore they should be less predictive of student success. A thorough variable importance analysis will be performed in order to test this assumption. 

\bigskip 

Understanding which predictors are important can also provide university administrators with feedback. For example, some of the High-Impact Practices identified by Randall Johnson and King Stage \citeyear{Johnson18} are part of the University of Toronto's program. The variable importance analysis could be a useful tool to assess the effect of such practices. 

\section{Methodology}
\label{sec:3}

\subsection{Data}
\label{data}

\bigskip

The data set provided by the University of Toronto contains 1 656 977 data points, where each observation represents the grade of one student in one course. A data point is a 7 dimensions observation containing the student ID, the course title, the department of the course, the semester, the credit value of the course and finally the numerical grade obtained by the student. As this is the only data obtained, some pre-processing is required in order for algorithms to be trained. The {\bf first research question} is whether it is possible to design an algorithm which accurately predicts whether or not a student will complete their program.  The {\bf second research question} is whether it is possible to design an algorithm which accurately predicts, for students who complete their program, which major they will complete. These two predictions will be based upon first-year student records.

\bigskip

The data has been pre-processed for the needs of the analyses. At the University of Toronto, a student must complete 20 credits in order to obtain an Honours B.A. or B.Sc \cite{UofT2017}. A student must also either complete 1 Specialist, 2 Majors or 1 Major and 2 Minors. The first five credits attempted by a student roughly represent one year of courses. Therefore, for each student every semester until the student reaches 5 attempted credits are used for prediction. It means that for some students, the predictors represent exactly 5 attempted credits and for some other students, a bit more. The set of predictors consists of the number of credits a student attempted in every department and the average grade across all courses taken by the student in each department. Since courses were taken by students in 71 different departments, the predictor vector is of length 142. Of course, many other predictors could also be computed from the data set, but these are the most appropriate ones for the purpose of the variable importance analysis. 

\bigskip

To answer the first research question, a binary response indicating whether or not a student completed their program is needed. Students that completed 18 credits were labelled as students who completed their program. Students who registered to 5 credits worth of courses, succeeded at fewer than 18 credits worth of courses and stopped taking courses for 3 consecutive semesters are considered students who began a program but did not complete it. All other students were left out of the analysis. Since some students take classes in other faculties or universities, 18 credits was deemed a reasonable threshold. It is possible that some students did not complete their program even though they completed 18 credits, but it is more likely that they took courses in other faculties or universities. To be considered dropouts, only students who registered to at least 5 credits worth of courses were considered. It was assumed that students that registered to fewer credits were registered in another faculty, campus, university or were simply auditing students.  After this pre-processing was performed, the data set contains 38 842 students of which 26 488 completed an undergraduate program and 12 294 did not.  

\bigskip

To answer the second research question a categorical response representing the major completed by the student is required. To do so, the 26 448 students who completed a program are kept. The response will represent the major completed by the student. Since this information is not available in the data set, the department in which the student completed the largest number of credits is considered the program they majored in. Therefore, the response variable is a categorical variable that can take 71 possible values. This formatting choice might be a problem for students who completed more than 1 major. Some recommendations to fix that problem can be found in the conclusion.  

\bigskip

Regarding the various grading policies of this university it was noticed that Mathematics, Chemistry and Economics are the three departments with the lowest average grades. As grades do vary widely across the data set there is no statistically significant difference between the departments but it is still interesting to observe that departments that were defined as low-grading departments in many papers do appear as the lowest grading departments in this data set too. Finally, the data set was divided in three parts as is it usually done. The algorithm is trained upon the training set, which contains 90\% of the observations in order to learn from a large portion of the data set. 5\% of the data set is assigned to the validation set which is utilized to select various optimization parameters. Finally, the rest of the data set is assigned to the test set, which is a data set totally left aside during training and later used to test the performances of the trained classifier.

\subsection{Classification Tree} 
\label{sectree}

\bigskip

A typical supervised statistical learning problem is defined when the relationship between a response variable and an associated set of predictors (used interchangeably with inputs) is of interest. The response is what needs prediction, such as the program completion, and the predictors, such as the grades, are used to predict the response.  When the response variable is categorical, this problem is defined as a classification problem. One challenge in classification problems is to use a data set in order to construct a classifier. A classifier is built to emit a class prediction for any new observation with unknown response. In this analysis, classifiers are built upon the data set described in section \ref{data} to predict if a new student will complete its program and what major will be completed using information related to its first year of courses. 

\bigskip

\begin{figure*}[ht]
\begin{center}
\includegraphics[width=6cm]{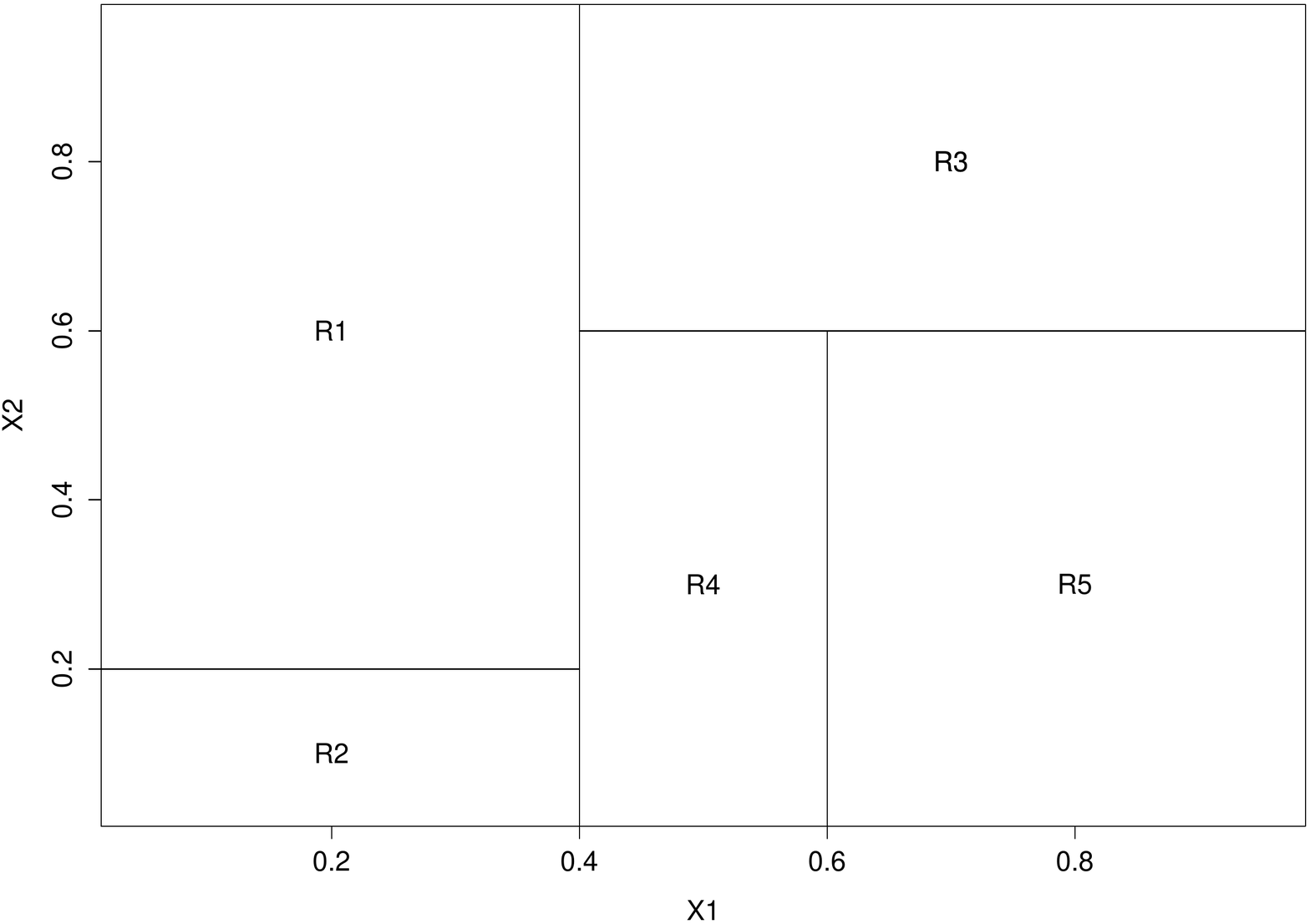}
\includegraphics[width=6cm]{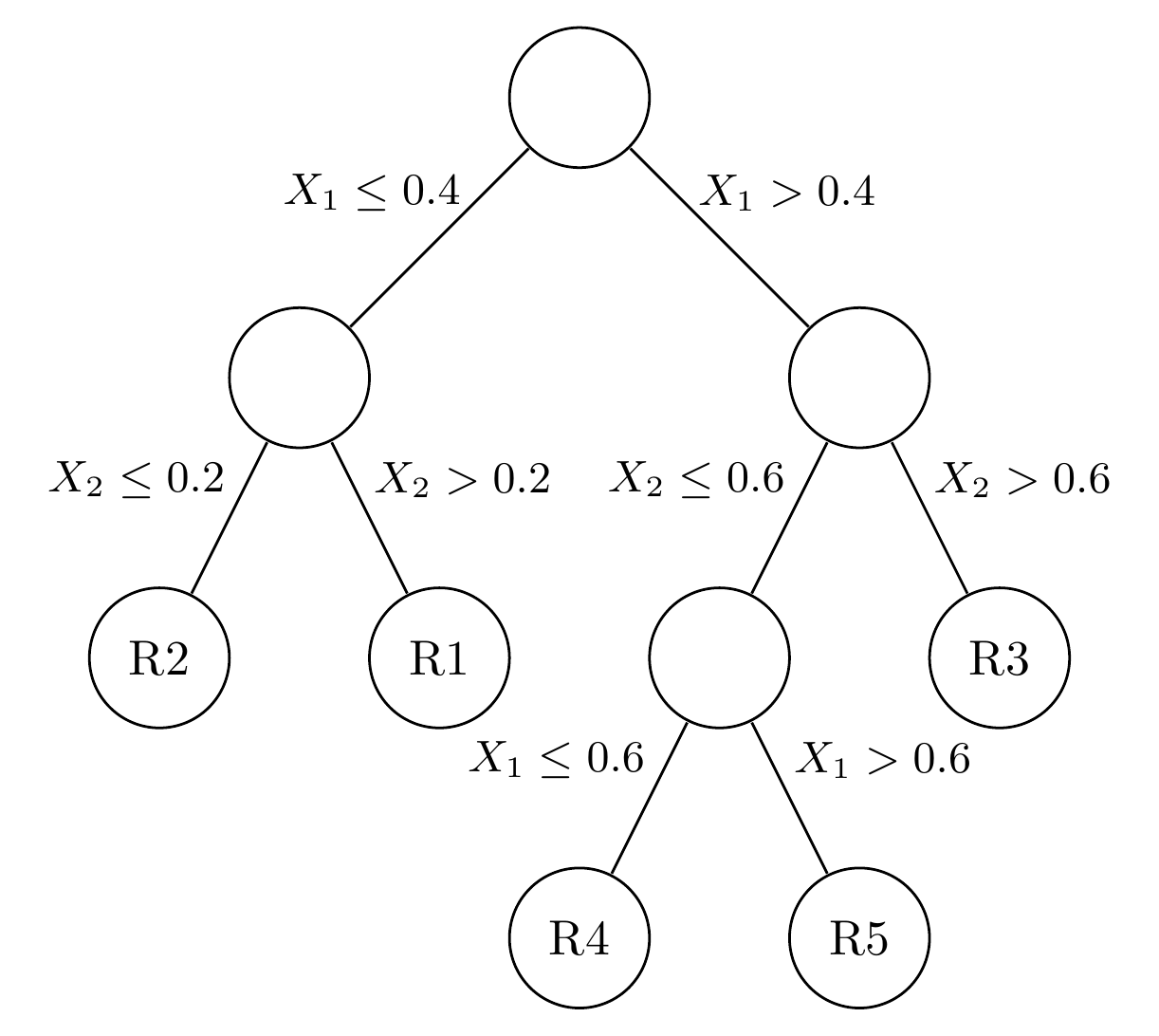}
\caption{Illustration of a decision tree partition of the predictor space  in 5 regions and the associated decision tree}
\label{figtree} 
\end{center}
\end{figure*}

A classification tree \cite{Breiman84} is a model that classifies new observations based on set of conditions related to the predictors. For example, a classification tree could predict a student is on its way to complete a program because it attempted more than 2 Mathematics courses, obtained an averaged grade in Mathematics above 80 and attempted fewer than 2 Psychology courses. The set of conditions established by a decision tree partitions in multiple regions the space defined by possible predictors values. Intuitively, a classification tree forms regions defined by some predictors values and assign a response label for new observations that would belong in those regions. Figure \ref{figtree} illustrates an example of a predictor space partition, its associated regions and its associated classification tree for observations defined by two predictors. The final set of regions can be defined as leaves in a tree as represented in Figure \ref{figtree}, hence the name classification trees.

\bigskip

Now that the model has been established, an algorithm that creates the classification tree using a training set of labelled observations needs to be defined. The algorithm creates the regions by recursively establishing the conditions. It aims at building regions that contains a high concentration of observations of the same class. Usually a measure of impurity is defined; the further the region is from containing only observations with the same label, the bigger this measure is. Intuitively, it is desired to obtain a set of conditions under which all students either completed their programs or not. Therefore, the algorithm analyses how mixed are the labels according to all possible conditions and selects the condition that minimizes the measure of impurity. For example, the algorithm will look at all conditions of the form : "did the student attempt more or less than 1 Mathematics course ?" and select the condition that best divides students that completed a program from students that did not. 

\bigskip

Once a condition is selected, the training observations are effectively divided in two sets of training observations based upon the condition. The process is repeatedly applied on the two resulting training sets. The algorithm divides the training observations in smaller sets until each resulting set contains few observations. When the partitioning process is completed, each region is labelled with the class representing the majority of observations respecting the conditions defining the region. A more formal definition of the algorithm is included in the appendix.

\subsection{Random Forest}
\label{secforest}

\bigskip

By constructing a decision tree, a powerful and easy to interpret classifier is obtained. As will be demonstrated in this section, one way to improve this classifier is to build a set of classifiers using samples of the training set.

\bigskip

Suppose there is a way to obtain a set of classifiers. The goal is to find a technique that uses the entire set of classifiers to get a new classifier that is better than any of them individually.  One method of aggregating the class predictions is by \textit{voting}: the predicted class for a new observation is the most picked class among individual classifier. A critical factor in whether the aggregating procedure will improve the accuracy or not is the stability of the individual classifiers. If a small variation in the training set has almost no effect on the classifier, this classifier is said to be stable, and utilizing a set of classifiers based upon similar training sets will result in a set of almost identical classifiers. For unstable procedures, the classifiers in the set are going to be very different from one another. For such classifiers, the aggregation will greatly improve both the stability and accuracy of the procedure. Procedure stability was studied by Breiman \citeyear{Breiman96a}; classification trees are unstable. 

\bigskip

Bootstrap aggregating (\textit{bagging}) was introduced by Breiman \citeyear{Breiman96} as a way to improve unstable classifiers. In bagging, each classifier in the set is built upon a different bootstrap sample of the training set. A bootstrap sample is simply a random sample of the original training sets. Each of the samples are drawn at random with replacement from the original training set and are of the same size. Doing so will produce a set of different training sets. For each of these training set a decision tree is fitted and together they form a random forest. Overfitting is a problem caused when a classifier identifies a structure that corresponds too closely to the training set and generalizes poorly to new observations. By generating multiple training sets, fitting multiple trees and building a forest out of these tree classifiers it greatly reduces the chances of overfitting. Breiman \citeyear{Breiman01} defines a \textit{random forest} as a classifier consisting of a set of tree-structured classifiers where each tree casts a unit vote for the most popular class at one input.

\bigskip

Breiman introduced in 2001 random forests with random inputs \cite{Breiman01} which is the most commonly used random forest classifier. The novelty of this random forest model is in the tree-growing procedure. Instead of finding the best condition among all the predictors, the algorithm will now randomly select a subset of predictors and will find the best condition among these, this modification greatly improved the accuracy of random forests.

\bigskip

Random forests are easy to use and are stable classifiers with many interesting properties. One of these interesting properties is that they allow for powerful variable importance computations that evaluate the importance of individual predictors throughout the entire prediction process.

\subsection{Variable Importance in Random Forests} \label{VISec}

A variable importance analysis aims at understanding the effect of individual predictors on the classifier output. A predictor with a great effect is considered an important predictor. A random forest provides multiple interesting variable importance computations. The \textit{Gini decrease importance} sums the total impurity measure decrease caused by partitioning upon a predictor throughout an entire tree and then computes the average of this measure across all trees in a forest. This technique is tightly related to the construction process of the tree itself and is pretty easy to obtain as it is non-demanding computationally. 

\bigskip

The \textit{permutation decrease importance} was introduced by Breiman \citeyear{Breiman01}. Intuitively if a predictor has a significant effect on the response, the algorithm should lose a lot of prediction accuracy if the values of that predictor are mixed up in the data set. One way to disrupt the predictors values is by permutations. The procedure computes the prediction accuracy on the test set using the true test set. Then, it permutes the values of one predictor, $j$, across all observations, run this permuted data through the forest and compute the new accuracy. If the input $j$ is important, the algorithm should lose a lot of its prediction accuracy by permuting the values of $j$ in the test set. The process is repeated for all predictors, then it is averaged across all trees and the averaged prediction accuracy decreases are compared. The larger the decrease in accuracy the more important the variable is considered.   

\bigskip

Storbl \& al. \citeyear{Strobl07} recently published an article where these techniques are analysed and compared. According to this paper, the selection bias of the decision tree procedure might lead to misleading variable importance. Numerous papers \cite{Breiman84,Loh01,Kononenko95} noticed a selection bias within the decision tree procedure when the predictors are of different nature.  The simulation studies produced by Storbl \& al. \citeyear{Strobl07} show that the Gini decrease importance is not a reliable variable importance measure when predictors are of varying types. The Gini decrease importance measure tends to overestimate the importance of continuous variables.

\bigskip

It is also shown \cite{Strobl07} that the variable importance techniques described above can give misleading results due the replacements when drawing bootstrap samples. It is recommended that researchers build random forests with bootstrap samples without replacements and use an unbiased tree-building procedure \cite{Loh97,Loh01,Loh02,Hothorn12}. If a classic tree-building procedure is used, predictors should be of the same type or only the permutation decrease importance is reliable.

\subsection{Algorithms}

\bigskip

A classification tree using the Gini impurity as split measurement was coded in the C++ language using the Rcpp library \cite{Eddelbuettel11}. The code is available upon request from the first author.  The algorithm proceeds as explained in Section \ref{sectree}, the tree it produces is unpruned and training sets are partitioned until they contain only 50 observations.  Three versions of the random forest algorithm are going to be used. Even though one of these models will outperform to two other in terms of prediction accuracy, the variable importance analysis of all three models will be considered and aggregate. For clarity and conciseness purposes, only the best model's performance will be assessed. \textbf{ Random forest \# 1} consists of 200 trees and can split upon every variable in each region. Bootstrap samples are drawn without replacement and contain 63\% of the original training set. \textbf{Random forest \# 2} fits 200 trees but randomly selects the variable to be partitioned upon in each region. 

\bigskip

Finally, the popular R RandomForest package \cite{Liaw02} was also used.  It is an easy to use and reliable package that can fit random forests and produce variable importance plots. Using this package, \textbf{random forest \# 3} was built. It contains 200 trees. Once again, bootstrap samples are drawn without replacement and contain about 63\% of the size of the original training set. By default, this algorithm randomly selects a subset of inputs for each region. Regarding the impurity measure, the Gini impurity was selected because it has interesting theoretical properties, such as being differentiable, and has been performing well empirically.  

\bigskip

Linear models were trained for both of the classification problems serving as benchmarks. In order for the comparison to be as direct as possible, the linear model classifiers were constructed upon the same set of predictors; it may be possible to improve both the random forest and the linear model with different predictors. As the problems are two classification ones, the linear models selected were logistic regression models and details regarding their parametrizations are included in the appendix.

\bigskip

\section{Results} \label{secres}

\subsection{First research question : Predicting program completion}

\bigskip

\textbf{Random forest \# 3} produced the best accuracy on the test set. Among the students who completed their program in the test set, the classifier achieves a 91.19\% accuracy. Out of the 418 students who did not complete their program, the classifier achieves a 52.95\% accuracy. The combined result it a 78.84\% accuracy over the complete test set. 

\bigskip

Obviously this is higher accuracy than if all students would be classified as students who competed their program, which would result in a 68.08\% accuracy. The random forest accuracy is also slightly higher than the 74.21\% accuracy achieved with a logistic regression based upon the same predictors. These predictions can be useful for university administrations that would like to predict the number of second-year students and prepare accordingly with a sufficient margin. About 75\% of students identified as dropouts by the random forest classifier are true dropouts. Therefore students identified as dropouts by the algorithm could be considered higher-risk students and these predictions could be useful in order to target students in need of more support to succeed. The relatively high accuracy of the classifier is also an indicator that the variable importance analysis is reliable.

\bigskip 

\begin{figure}[!htb]
\begin{center}
\includegraphics[width=14cm,height=10cm]{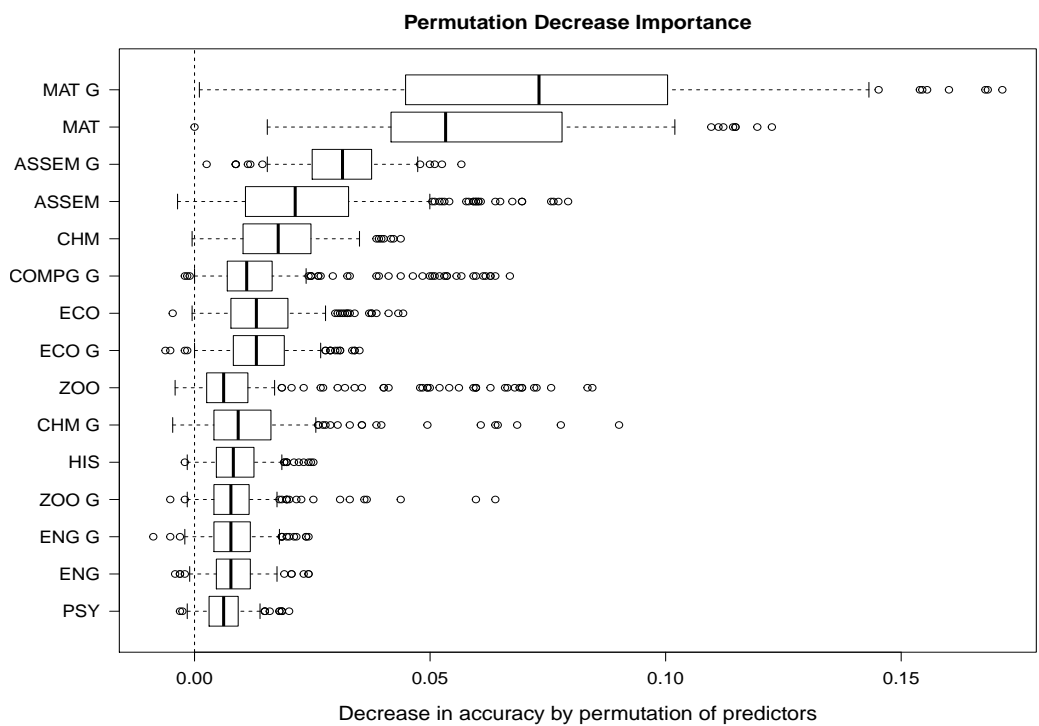}
\caption{Variables importance boxplots for the \textbf{random forest \# 1}.}
\label{FoN_VI_RF}
\end{center} 
\end{figure}

\begin{figure}[!htb]
\begin{center}
\includegraphics[width=14cm,height=10cm]{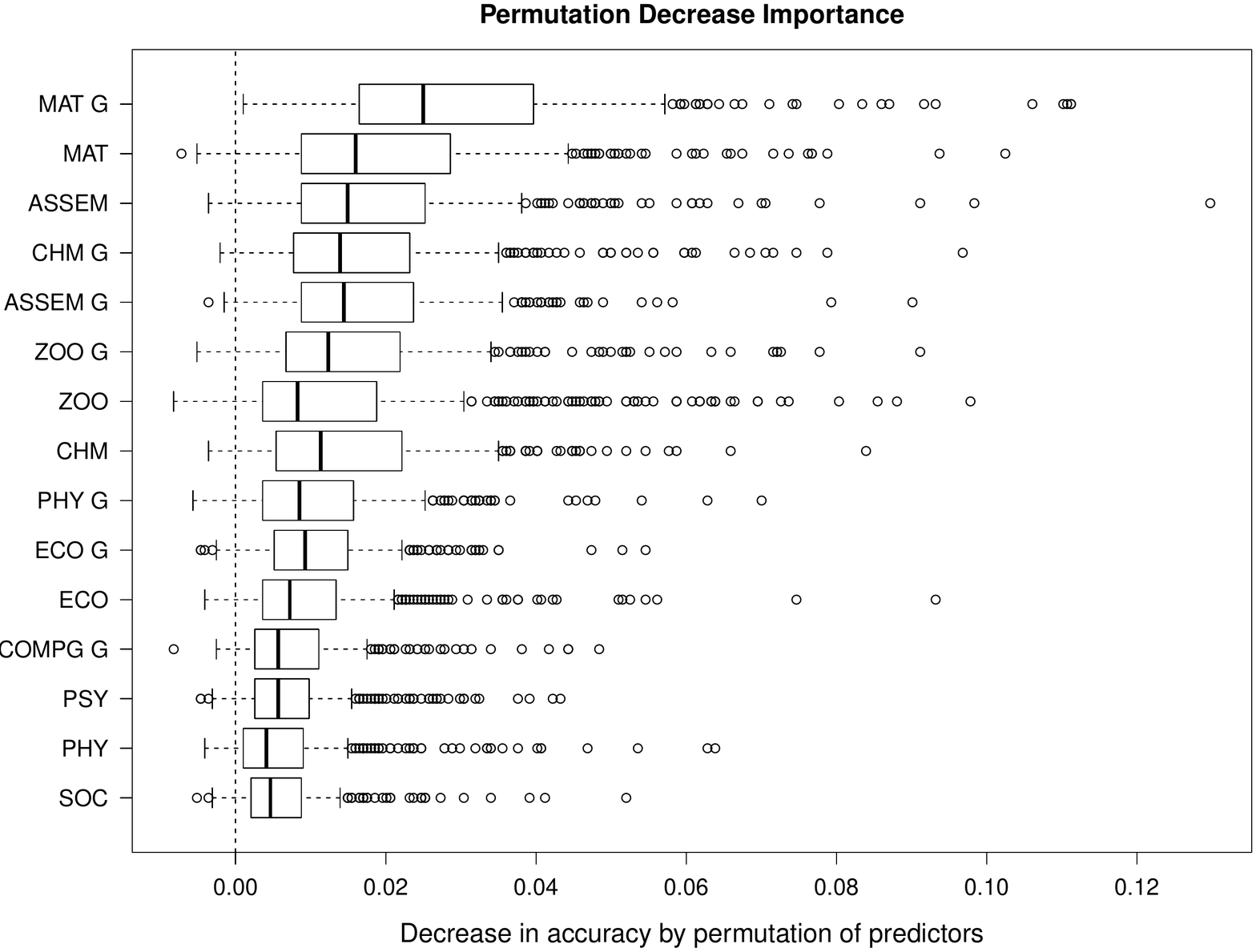}
\caption{Variables importance boxplots for the \textbf{random forest \# 2}.}
\label{FoN_VI_RFRI}
\end{center}
\end{figure}

\begin{figure}[!htb]
\begin{center}
\includegraphics[width=14cm,height=10cm]{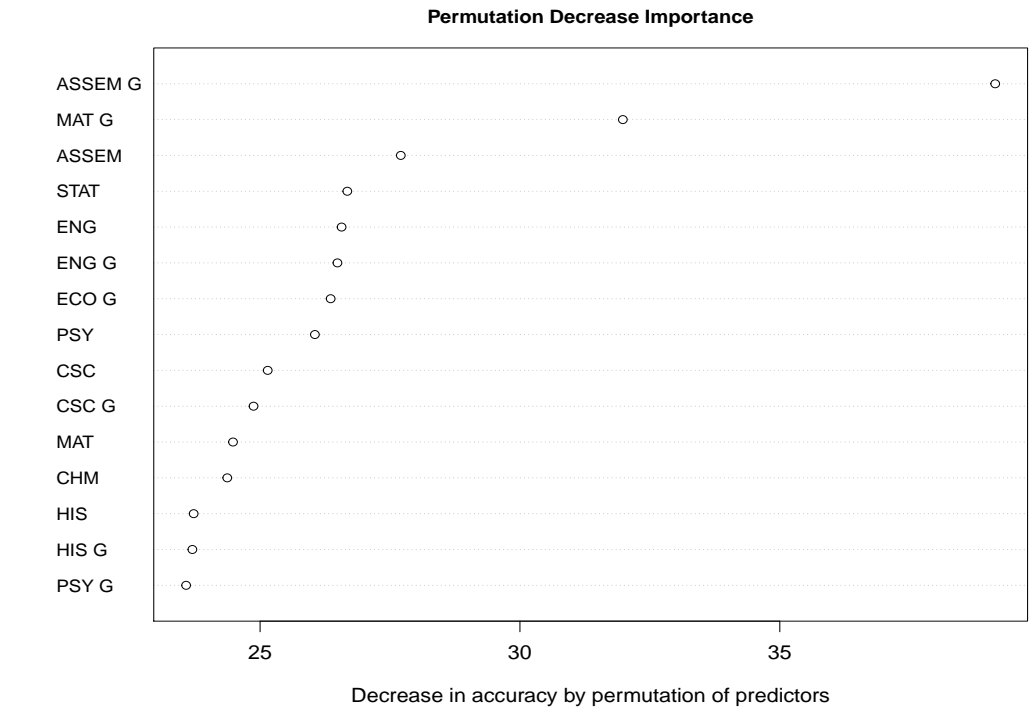}
\caption{Variable importance plot produced by the RandomForest package for the \textbf{random forest \# 3}.}
\label{FoN_VI_RFPack} 
\end{center}
\end{figure}

Variable importance is determined by the average decrease in accuracy in the test set caused by a random permutation of the predictor. This technique has been selected since it is more reliable as explained in Section \ref{VISec}. The top 15 variables according to the permutation decrease were kept and ordered in Figures \ref{FoN_VI_RF},\ref{FoN_VI_RFRI} and \ref{FoN_VI_RFPack}. Since variable importance varies from one model to another, the three variable importance plots were included and the results will be aggregated.

In Figures \ref{FoN_VI_RF},\ref{FoN_VI_RFRI} and \ref{FoN_VI_RFPack} and for all the following figures, the variable representing the number of credits in a department is identified by the department code, i.e. the number of credits in Chemistry is identified by CHM. The variable representing the averaged grade in a department is identified by the department code followed by the letter G, i.e CHM G represents the averaged grade in Chemistry. 

\bigskip

To begin, it was also noted that the variance for the grade variables were larger. Across all three random forests, the grades in Mathematics (MAT), Finance (COMPG), Economics (ECO) are consistently among the most important grade variable. These departments are considered low-grading departments and perhaps the strict marking of these departments helps to better distinguish students among themselves. A possible explanation is that the grade inflation that suffered the high-grading departments caused the grades to be no longer a reliable tool to distinguish students among themselves which could be a symptom of grade inflation as suggested in section \ref{gi}. Other factors could have caused this phenomenon such as less sequential courses in Human Science fields, larger classes size, reduced access to a professor or other factors. It is impossible to claim for sure that these results are caused by the grade inflation problem, but these results could indicate such thing. Therefore, universities could use such technique to verify if grades in a department have more predictive power than grades in other departments and act accordingly since grades should represent students' abilities. 

\bigskip

It is also important to notice the importance of ASSEM in the three variable importance plots. The ASSEM code represents a special type of first year seminar course. It seems that the students that registers in theses courses are easy to classify as both grades and the number of credits are considered important. This result agrees with the result obtained by Johnson and Stage \citeyear{Johnson18} about the importance of first year seminar courses. The first year seminar courses (ASSEM) were brand new at the University of Toronto and the analysis performed provided evidence of the merit of such courses in order to establish a student's profile and to predict success. In other words, such variable importance analysis could help university administrations assess the usefulness of new programs and courses.   

\bigskip

\subsection{Second research question : Predicting the major}

\bigskip

The second task at hand is to build a random forest that predicts the student's major. Once again, from a prediction accuracy perspective, \textbf{random forest \# 3} offered better performances with a 47.41\% accuracy in predicting the major completed. This appears slightly lower than expected, but considering there are 71 different programs, being able to pin down the right program for about half of the students seems successful. This is a better result than the meager 4.75\% obtained by assigning majors with probabilities weighted by the proportion of the majors completed. The 47.41\% accuracy of the random forest is also above the 42.63\% accuracy obtained by the multinomial logistic regression benchmark. For classification purposes, these classifiers could help individual departments predict the number of students registering to second, third or fourth year courses and graduate programs. Predicting the major could also help university administrations to allocate the financial resources among the departments or to decide the programs that require more advertisements.

\bigskip

Variable importance is also interesting for that research questions. Here is the variable importance analyses produced by the three random forests; once again, the 15 most important predictors are displayed. The importance of a predictor is determined by the average decrease in accuracy in the test set caused by a random permutation of the predictor.
  
\bigskip

\begin{figure}[!htb]
\begin{center}
\includegraphics[width=14cm,height=10cm]{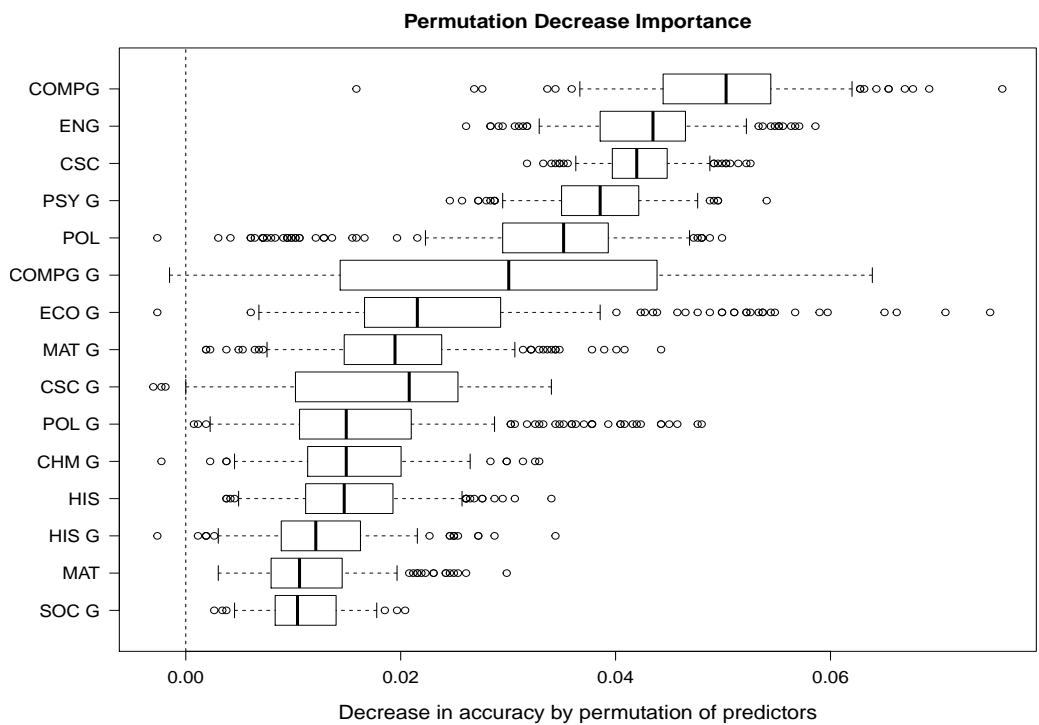}
\caption{Variables importance boxplots for the \textbf{random forest \# 1}. }
\label{MC_VI_RF} 
\end{center}
\end{figure}

\begin{figure}[!htb]
\begin{center}
\includegraphics[width=14cm,height=10cm]{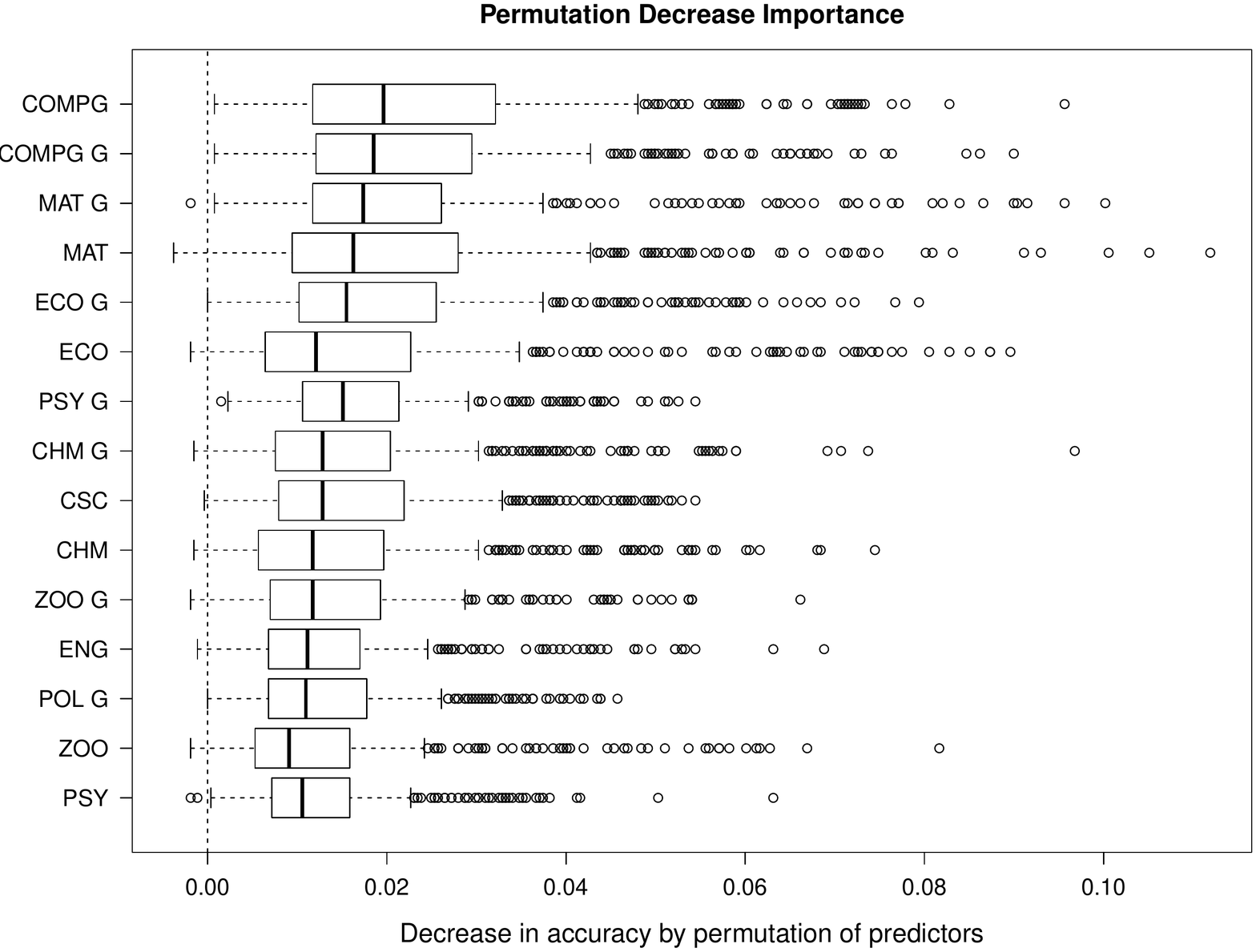}
\caption{Variables importance boxplots for the \textbf{random forest \# 2}. }
\label{MC_VI_RFRI} 
\end{center}
\end{figure}

\begin{figure}[!htb]
\begin{center}
\includegraphics[width=14cm,height=10cm]{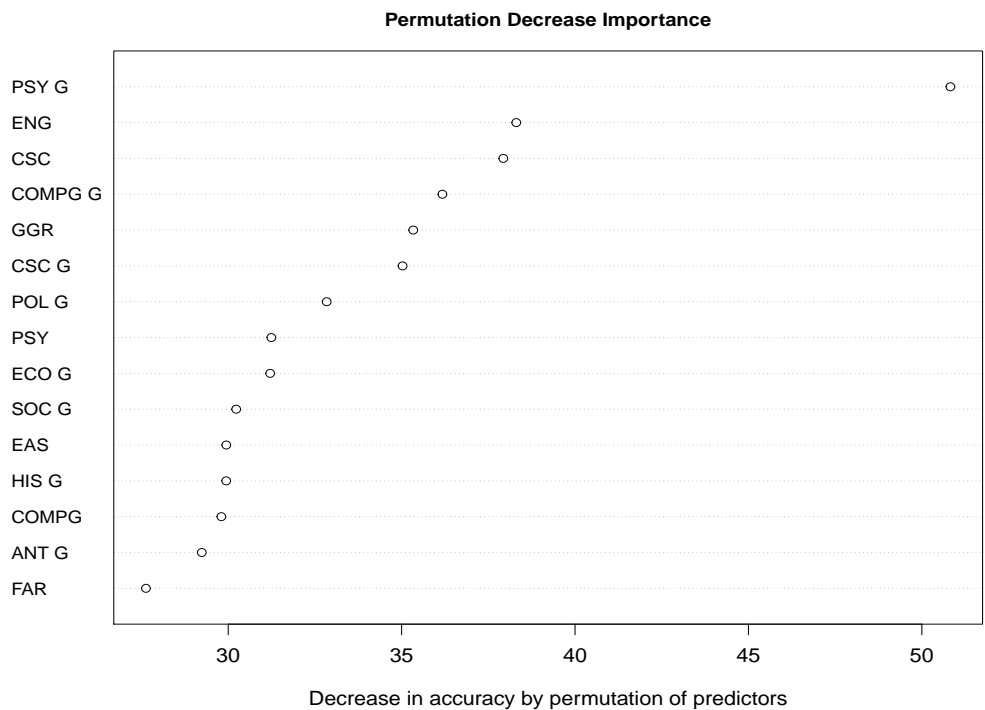}
\caption{Variable importance plot produced by the RandomForest package for the \textbf{random forest \# 3}.}
\label{MC_VI_RFPack} 
\end{center}
\end{figure}

\bigskip

A decrease in importance for the grades variable is noted in Figure \ref{MC_VI_RF},\ref{MC_VI_RFRI} and \ref{MC_VI_RFPack}. This was to be expected because of how the data was formatted. Since the department in which the highest amount of credit was obtained is considered the major completed by the student, these variable importance measures are not surprising. Actually, if all the courses were included, instead of only the first year, the amount of credit in every department precisely defines the response variable. Considering this weakness in the data formatting, the grades still have a relatively high importance. It seems hard to see any effect of grading policies in the predictive power of grades regarding that research question.

\bigskip

It seems like for some departments, such as English (ENG) and Computers Sciences (CSC), it is easy to predict students that will complete a major in those departments by almost solely looking at the number of courses attempted in those departments during the first year. This is caused by the fact that a vast majority of students that take courses in Computers Science or English during their first year end up completing an undergraduate program in these departments respectively. From a policy-making perspective, departments could use this information as they might want to adapt the content of their first-year courses now that they know more about the audience of these courses. 

\section{Conclusion}

The first year's worth of courses and grades were used to build two classifiers; one that predicts if a student will complete their undergraduate program, the other that predicts the major of a student who completed a program. Random forests were used to build those classifiers. Random forests are easy to use with most statistical computing languages, fast to train, and they outperform linear logistic models in terms of prediction accuracy. For practitioners, random forests could be an alternative to typical linear models for various prediction tasks; to predict the number of students registered in second-year courses, the distribution of students across the many programs or to identify students at risk of failing or dropping out.

\bigskip

Evaluating the importance of each predictor is also something that offers random forest in comparison to the benchmark model. In this study, it was observed in Section \ref{secres} that grades were important for predicting if a student will complete their program. Grades in departments that were considered low-grading departments in some grades inflation research articles like Mathematics, Economics and Finance are consistently among the most important variables.  These results indicate that a strong relationship exists between the grades in low-grading departments and the chance of succeeding at an undergraduate program, although this does not necessarily indicate a {\it causal} connection. Grades were somewhat less important predictors for predicting the students' major but even though they were less important, grades in Mathematics, Finance, Economics and Psychology (PSY) were still frequently significantly important. 

\bigskip

Finally, for potential improvements in the data analysis, it is to be noted that some students might have completed more than one major or specialization. This might explain the relatively low accuracy for major choice prediction. Allowing for multiple major choices is a potential improvement for this model. This is in fact a multi-label classification problem and some solutions have already been proposed to adapt decision trees to accommodate this more complicated problem \cite{Clare01,Chen03,Chou05}. Some departments also share a great deal of similarities and might be considered equivalent by the university, thus combining some of them might increase the prediction accuracy. The missing values in the predictors were also problematic. Ideally, the algorithm would consider splitting on the grade variables for a certain department only to classify students who took courses in that department. Developing a new decision tree algorithm where new variables are added to the pool of potential split variables depending on previous partitioning should be a great way to improve the actual model in certain scenarios. Overall, implementing a new tree-building procedure where variable are added or discarded based upon previous partitioning and considering a multi-label classifier like suggested by Chen \& al. \citeyear{Chen03} could be great improvements for future work on that data set. 

\section*{Acknowledgement}
We are very grateful to Glenn Loney and Sinisa Markovic of the University of Toronto for providing us with students grade data. The authors also gratefully acknowledge the financial support from the NSERC of Canada.

\bibliographystyle{apacite}     
\bibliography{mybibfile}   

\begin{thebibliography}{}

\bibitem [\protect \citeauthoryear {%
{Aulck}%
, {Velagapudi}%
, {Blumenstock}%
\BCBL {}\ \BBA {} {West}%
}{%
{Aulck}%
\ \protect \BOthers {.}}{%
{\protect \APACyear {2016}}%
}]{%
Aulck16}
\APACinsertmetastar {%
Aulck16}%
\begin{APACrefauthors}%
{Aulck}, L.%
, {Velagapudi}, N.%
, {Blumenstock}, J.%
\BCBL {}\ \BBA {} {West}, J.%
\end{APACrefauthors}%
\unskip\
\newblock
\APACrefYearMonthDay{2016}{{\APACmonth{06}}}{}.
\newblock
{\BBOQ}\APACrefatitle {{Predicting Student Dropout in Higher Education}}
  {{Predicting Student Dropout in Higher Education}}.{\BBCQ}
\newblock
\APACjournalVolNumPages{ArXiv e-prints}{}{}{}.
\PrintBackRefs{\CurrentBib}

\bibitem [\protect \citeauthoryear {%
Bailey%
, Rosenthal%
\BCBL {}\ \BBA {} Yoon%
}{%
Bailey%
\ \protect \BOthers {.}}{%
{\protect \APACyear {2016}}%
}]{%
Bailey16}
\APACinsertmetastar {%
Bailey16}%
\begin{APACrefauthors}%
Bailey, M\BPBI A.%
, Rosenthal, J\BPBI S.%
\BCBL {}\ \BBA {} Yoon, A\BPBI H.%
\end{APACrefauthors}%
\unskip\
\newblock
\APACrefYearMonthDay{2016}{}{}.
\newblock
{\BBOQ}\APACrefatitle {Grades and incentives: assessing competing grade point
  average measures and postgraduate outcomes} {Grades and incentives: assessing
  competing grade point average measures and postgraduate outcomes}.{\BBCQ}
\newblock
\APACjournalVolNumPages{Studies in Higher Education}{41}{9}{1548-1562}.
\newblock
\begin{APACrefURL} \url{http://dx.doi.org/10.1080/03075079.2014.982528}
  \end{APACrefURL}
\newblock
\begin{APACrefDOI} \doi{10.1080/03075079.2014.982528} \end{APACrefDOI}
\PrintBackRefs{\CurrentBib}

\bibitem [\protect \citeauthoryear {%
Bar%
, Kadiyali%
\BCBL {}\ \BBA {} Zussman%
}{%
Bar%
\ \protect \BOthers {.}}{%
{\protect \APACyear {2009}}%
}]{%
Bar09}
\APACinsertmetastar {%
Bar09}%
\begin{APACrefauthors}%
Bar, T.%
, Kadiyali, V.%
\BCBL {}\ \BBA {} Zussman, A.%
\end{APACrefauthors}%
\unskip\
\newblock
\APACrefYearMonthDay{2009}{}{}.
\newblock
{\BBOQ}\APACrefatitle {Grade Information and Grade Inflation: The Cornell
  Experiment} {Grade information and grade inflation: The cornell
  experiment}.{\BBCQ}
\newblock
\APACjournalVolNumPages{Journal of Economic Perspectivs}{23}{3}{93--108}.
\PrintBackRefs{\CurrentBib}

\bibitem [\protect \citeauthoryear {%
Breiman%
}{%
Breiman%
}{%
{\protect \APACyear {1996}}%
{\protect \APACexlab {{\protect \BCnt {1}}}}}]{%
Breiman96}
\APACinsertmetastar {%
Breiman96}%
\begin{APACrefauthors}%
Breiman, L.%
\end{APACrefauthors}%
\unskip\
\newblock
\APACrefYearMonthDay{1996{\protect \BCnt {1}}}{}{}.
\newblock
{\BBOQ}\APACrefatitle {Bagging predictors} {Bagging predictors}.{\BBCQ}
\newblock
\APACjournalVolNumPages{Machine Learning}{24}{2}{123--140}.
\newblock
\begin{APACrefURL} \url{http://dx.doi.org/10.1007/BF00058655} \end{APACrefURL}
\newblock
\begin{APACrefDOI} \doi{10.1007/BF00058655} \end{APACrefDOI}
\PrintBackRefs{\CurrentBib}

\bibitem [\protect \citeauthoryear {%
Breiman%
}{%
Breiman%
}{%
{\protect \APACyear {1996}}%
{\protect \APACexlab {{\protect \BCnt {2}}}}}]{%
Breiman96a}
\APACinsertmetastar {%
Breiman96a}%
\begin{APACrefauthors}%
Breiman, L.%
\end{APACrefauthors}%
\unskip\
\newblock
\APACrefYearMonthDay{1996{\protect \BCnt {2}}}{12}{}.
\newblock
{\BBOQ}\APACrefatitle {Heuristics of instability and stabilization in model
  selection} {Heuristics of instability and stabilization in model
  selection}.{\BBCQ}
\newblock
\APACjournalVolNumPages{Ann. Statist.}{24}{6}{2350--2383}.
\newblock
\begin{APACrefURL} \url{http://dx.doi.org/10.1214/aos/1032181158}
  \end{APACrefURL}
\newblock
\begin{APACrefDOI} \doi{10.1214/aos/1032181158} \end{APACrefDOI}
\PrintBackRefs{\CurrentBib}

\bibitem [\protect \citeauthoryear {%
Breiman%
}{%
Breiman%
}{%
{\protect \APACyear {2001}}%
}]{%
Breiman01}
\APACinsertmetastar {%
Breiman01}%
\begin{APACrefauthors}%
Breiman, L.%
\end{APACrefauthors}%
\unskip\
\newblock
\APACrefYearMonthDay{2001}{}{}.
\newblock
{\BBOQ}\APACrefatitle {Random Forests} {Random forests}.{\BBCQ}
\newblock
\APACjournalVolNumPages{Machine Learning}{45}{1}{5--32}.
\newblock
\begin{APACrefURL} \url{http://dx.doi.org/10.1023/A:1010933404324}
  \end{APACrefURL}
\newblock
\begin{APACrefDOI} \doi{10.1023/A:1010933404324} \end{APACrefDOI}
\PrintBackRefs{\CurrentBib}

\bibitem [\protect \citeauthoryear {%
{Breiman}%
, {Friedman}%
, {Olshen}%
\BCBL {}\ \BBA {} {Stone}%
}{%
{Breiman}%
\ \protect \BOthers {.}}{%
{\protect \APACyear {1984}}%
}]{%
Breiman84}
\APACinsertmetastar {%
Breiman84}%
\begin{APACrefauthors}%
{Breiman}, L.%
, {Friedman}, J\BPBI H.%
, {Olshen}, R\BPBI A.%
\BCBL {}\ \BBA {} {Stone}, C\BPBI J.%
\end{APACrefauthors}%
\unskip\
\newblock
\APACrefYear{1984}.
\newblock
\APACrefbtitle {Classification and Regression Trees} {Classification and
  regression trees}.
\newblock
\APACaddressPublisher{Belmont, California, U.S.A.}{Wadsworth Publishing
  Company}.
\PrintBackRefs{\CurrentBib}

\bibitem [\protect \citeauthoryear {%
R.~Chen%
\ \BBA {} DesJardins%
}{%
R.~Chen%
\ \BBA {} DesJardins%
}{%
{\protect \APACyear {2008}}%
}]{%
Chen08}
\APACinsertmetastar {%
Chen08}%
\begin{APACrefauthors}%
Chen, R.%
\BCBT {}\ \BBA {} DesJardins, S\BPBI L.%
\end{APACrefauthors}%
\unskip\
\newblock
\APACrefYearMonthDay{2008}{Feb}{01}.
\newblock
{\BBOQ}\APACrefatitle {Exploring the Effects of Financial Aid on the Gap in
  Student Dropout Risks by Income Level} {Exploring the effects of financial
  aid on the gap in student dropout risks by income level}.{\BBCQ}
\newblock
\APACjournalVolNumPages{Research in Higher Education}{49}{1}{1--18}.
\newblock
\begin{APACrefURL} \url{https://doi.org/10.1007/s11162-007-9060-9}
  \end{APACrefURL}
\newblock
\begin{APACrefDOI} \doi{10.1007/s11162-007-9060-9} \end{APACrefDOI}
\PrintBackRefs{\CurrentBib}

\bibitem [\protect \citeauthoryear {%
R.~Chen%
\ \BBA {} DesJardins%
}{%
R.~Chen%
\ \BBA {} DesJardins%
}{%
{\protect \APACyear {2010}}%
}]{%
Chen10}
\APACinsertmetastar {%
Chen10}%
\begin{APACrefauthors}%
Chen, R.%
\BCBT {}\ \BBA {} DesJardins, S\BPBI L.%
\end{APACrefauthors}%
\unskip\
\newblock
\APACrefYearMonthDay{2010}{}{}.
\newblock
{\BBOQ}\APACrefatitle {Investigating the Impact of Financial Aid on Student
  Dropout Risks: Racial and Ethnic Differences} {Investigating the impact of
  financial aid on student dropout risks: Racial and ethnic
  differences}.{\BBCQ}
\newblock
\APACjournalVolNumPages{The Journal of Higher Education}{81}{2}{179--208}.
\newblock
\begin{APACrefURL} \url{http://www.jstor.org/stable/40606850} \end{APACrefURL}
\PrintBackRefs{\CurrentBib}

\bibitem [\protect \citeauthoryear {%
Y\BHBI L.~Chen%
, Hsu%
\BCBL {}\ \BBA {} Chou%
}{%
Y\BHBI L.~Chen%
\ \protect \BOthers {.}}{%
{\protect \APACyear {2003}}%
}]{%
Chen03}
\APACinsertmetastar {%
Chen03}%
\begin{APACrefauthors}%
Chen, Y\BHBI L.%
, Hsu, C\BHBI L.%
\BCBL {}\ \BBA {} Chou, S\BHBI C.%
\end{APACrefauthors}%
\unskip\
\newblock
\APACrefYearMonthDay{2003}{}{}.
\newblock
{\BBOQ}\APACrefatitle {Constructing a multi-valued and multi-labeled decision
  tree} {Constructing a multi-valued and multi-labeled decision tree}.{\BBCQ}
\newblock
\APACjournalVolNumPages{Expert Systems with Applications}{25}{2}{199 - 209}.
\newblock
\begin{APACrefURL}
  \url{http://www.sciencedirect.com/science/article/pii/S0957417403000472}
  \end{APACrefURL}
\newblock
\begin{APACrefDOI} \doi{http://dx.doi.org/10.1016/S0957-4174(03)00047-2}
  \end{APACrefDOI}
\PrintBackRefs{\CurrentBib}

\bibitem [\protect \citeauthoryear {%
Chou%
\ \BBA {} Hsu%
}{%
Chou%
\ \BBA {} Hsu%
}{%
{\protect \APACyear {2005}}%
}]{%
Chou05}
\APACinsertmetastar {%
Chou05}%
\begin{APACrefauthors}%
Chou, S.%
\BCBT {}\ \BBA {} Hsu, C\BHBI L.%
\end{APACrefauthors}%
\unskip\
\newblock
\APACrefYearMonthDay{2005}{{\APACmonth{05}}}{}.
\newblock
{\BBOQ}\APACrefatitle {{MMDT}: A Multi-valued and Multi-labeled Decision Tree
  Classifier for Data Mining} {{MMDT}: A multi-valued and multi-labeled
  decision tree classifier for data mining}.{\BBCQ}
\newblock
\APACjournalVolNumPages{Expert Syst. Appl.}{28}{4}{799--812}.
\newblock
\begin{APACrefURL} \url{http://dx.doi.org/10.1016/j.eswa.2004.12.035}
  \end{APACrefURL}
\newblock
\begin{APACrefDOI} \doi{10.1016/j.eswa.2004.12.035} \end{APACrefDOI}
\PrintBackRefs{\CurrentBib}

\bibitem [\protect \citeauthoryear {%
Clare%
\ \BBA {} King%
}{%
Clare%
\ \BBA {} King%
}{%
{\protect \APACyear {2001}}%
}]{%
Clare01}
\APACinsertmetastar {%
Clare01}%
\begin{APACrefauthors}%
Clare, A.%
\BCBT {}\ \BBA {} King, R\BPBI D.%
\end{APACrefauthors}%
\unskip\
\newblock
\APACrefYearMonthDay{2001}{}{}.
\newblock
{\BBOQ}\APACrefatitle {Knowledge Discovery in Multi-label Phenotype Data}
  {Knowledge discovery in multi-label phenotype data}.{\BBCQ}
\newblock
\BIn{} L.~De~Raedt\ \BBA {} A.~Siebes\ (\BEDS), \APACrefbtitle {Principles of
  Data Mining and Knowledge Discovery: 5th European Conference, PKDD 2001,
  Freiburg, Germany, September 3--5, 2001 Proceedings} {Principles of data
  mining and knowledge discovery: 5th european conference, pkdd 2001, freiburg,
  germany, september 3--5, 2001 proceedings}\ (\BPGS\ 42--53).
\newblock
\APACaddressPublisher{Berlin, Heidelberg}{Springer Berlin Heidelberg}.
\newblock
\begin{APACrefURL} \url{http://dx.doi.org/10.1007/3-540-44794-6}
  \end{APACrefURL}
\newblock
\begin{APACrefDOI} \doi{10.1007/3-540-44794-6} \end{APACrefDOI}
\PrintBackRefs{\CurrentBib}

\bibitem [\protect \citeauthoryear {%
Eddelbuettel%
\ \BBA {} Francois%
}{%
Eddelbuettel%
\ \BBA {} Francois%
}{%
{\protect \APACyear {2011}}%
}]{%
Eddelbuettel11}
\APACinsertmetastar {%
Eddelbuettel11}%
\begin{APACrefauthors}%
Eddelbuettel, D.%
\BCBT {}\ \BBA {} Francois, R.%
\end{APACrefauthors}%
\unskip\
\newblock
\APACrefYearMonthDay{2011}{}{}.
\newblock
{\BBOQ}\APACrefatitle {Rcpp: Seamless {R} and {C}++ Integration} {Rcpp:
  Seamless {R} and {C}++ integration}.{\BBCQ}
\newblock
\APACjournalVolNumPages{Journal of Statistical Software}{40}{1}{1--18}.
\newblock
\begin{APACrefURL}
  \url{https://www.jstatsoft.org/index.php/jss/article/view/v040i08}
  \end{APACrefURL}
\newblock
\begin{APACrefDOI} \doi{10.18637/jss.v040.i08} \end{APACrefDOI}
\PrintBackRefs{\CurrentBib}

\bibitem [\protect \citeauthoryear {%
Glaesser%
\ \BBA {} Cooper%
}{%
Glaesser%
\ \BBA {} Cooper%
}{%
{\protect \APACyear {2012}}%
}]{%
Glaesser12}
\APACinsertmetastar {%
Glaesser12}%
\begin{APACrefauthors}%
Glaesser, J.%
\BCBT {}\ \BBA {} Cooper, B.%
\end{APACrefauthors}%
\unskip\
\newblock
\APACrefYearMonthDay{2012}{}{}.
\newblock
{\BBOQ}\APACrefatitle {Gender, parental education, and ability: their
  interacting roles in predicting GCSE success} {Gender, parental education,
  and ability: their interacting roles in predicting gcse success}.{\BBCQ}
\newblock
\APACjournalVolNumPages{Cambridge Journal of Education}{42}{4}{463-480}.
\newblock
\begin{APACrefURL} \url{https://doi.org/10.1080/0305764X.2012.733346}
  \end{APACrefURL}
\newblock
\begin{APACrefDOI} \doi{10.1080/0305764X.2012.733346} \end{APACrefDOI}
\PrintBackRefs{\CurrentBib}

\bibitem [\protect \citeauthoryear {%
Hastie%
, Tibshirani%
\BCBL {}\ \BBA {} Friedman%
}{%
Hastie%
\ \protect \BOthers {.}}{%
{\protect \APACyear {2009}}%
}]{%
Hastie09}
\APACinsertmetastar {%
Hastie09}%
\begin{APACrefauthors}%
Hastie, T.%
, Tibshirani, R.%
\BCBL {}\ \BBA {} Friedman, J.%
\end{APACrefauthors}%
\unskip\
\newblock
\APACrefYear{2009}.
\newblock
\APACrefbtitle {The Elements of Statistical Learning} {The elements of
  statistical learning}\ (\PrintOrdinal{2}\ \BEd).
\newblock
\APACaddressPublisher{}{Springer}.
\PrintBackRefs{\CurrentBib}

\bibitem [\protect \citeauthoryear {%
Hothorn%
, Hornik%
\BCBL {}\ \BBA {} Zeileis%
}{%
Hothorn%
\ \protect \BOthers {.}}{%
{\protect \APACyear {2006}}%
}]{%
Hothorn12}
\APACinsertmetastar {%
Hothorn12}%
\begin{APACrefauthors}%
Hothorn, T.%
, Hornik, K.%
\BCBL {}\ \BBA {} Zeileis, A.%
\end{APACrefauthors}%
\unskip\
\newblock
\APACrefYearMonthDay{2006}{}{}.
\newblock
{\BBOQ}\APACrefatitle {Unbiased Recursive Partitioning: A Conditional Inference
  Framework} {Unbiased recursive partitioning: A conditional inference
  framework}.{\BBCQ}
\newblock
\APACjournalVolNumPages{Journal of Computational and Graphical
  Statistics}{15}{3}{651-674}.
\newblock
\begin{APACrefURL} \url{http://dx.doi.org/10.1198/106186006X133933}
  \end{APACrefURL}
\newblock
\begin{APACrefDOI} \doi{10.1198/106186006X133933} \end{APACrefDOI}
\PrintBackRefs{\CurrentBib}

\bibitem [\protect \citeauthoryear {%
S\BPBI R.~Johnson%
\ \BBA {} Stage%
}{%
S\BPBI R.~Johnson%
\ \BBA {} Stage%
}{%
{\protect \APACyear {2018}}%
}]{%
Johnson18}
\APACinsertmetastar {%
Johnson18}%
\begin{APACrefauthors}%
Johnson, S\BPBI R.%
\BCBT {}\ \BBA {} Stage, F\BPBI K.%
\end{APACrefauthors}%
\unskip\
\newblock
\APACrefYearMonthDay{2018}{}{}.
\newblock
{\BBOQ}\APACrefatitle {Academic Engagement and Student Success: Do High-Impact
  Practices Mean Higher Graduation Rates?} {Academic engagement and student
  success: Do high-impact practices mean higher graduation rates?}{\BBCQ}
\newblock
\APACjournalVolNumPages{The Journal of Higher Education}{0}{0}{1-29}.
\newblock
\begin{APACrefURL} \url{https://doi.org/10.1080/00221546.2018.1441107}
  \end{APACrefURL}
\newblock
\begin{APACrefDOI} \doi{10.1080/00221546.2018.1441107} \end{APACrefDOI}
\PrintBackRefs{\CurrentBib}

\bibitem [\protect \citeauthoryear {%
V\BPBI E.~Johnson%
}{%
V\BPBI E.~Johnson%
}{%
{\protect \APACyear {2003}}%
}]{%
Johnson03}
\APACinsertmetastar {%
Johnson03}%
\begin{APACrefauthors}%
Johnson, V\BPBI E.%
\end{APACrefauthors}%
\unskip\
\newblock
\APACrefYear{2003}.
\newblock
\APACrefbtitle {Grade Inflation : A Crisis in College Education} {Grade
  inflation : A crisis in college education}.
\newblock
\APACaddressPublisher{}{Springer}.
\PrintBackRefs{\CurrentBib}

\bibitem [\protect \citeauthoryear {%
Kappe%
\ \BBA {} van~der Flier%
}{%
Kappe%
\ \BBA {} van~der Flier%
}{%
{\protect \APACyear {2012}}%
}]{%
Kappe12}
\APACinsertmetastar {%
Kappe12}%
\begin{APACrefauthors}%
Kappe, R.%
\BCBT {}\ \BBA {} van~der Flier, H.%
\end{APACrefauthors}%
\unskip\
\newblock
\APACrefYearMonthDay{2012}{Dec}{01}.
\newblock
{\BBOQ}\APACrefatitle {Predicting academic success in higher education: what's
  more important than being smart?} {Predicting academic success in higher
  education: what's more important than being smart?}{\BBCQ}
\newblock
\APACjournalVolNumPages{European Journal of Psychology of
  Education}{27}{4}{605--619}.
\newblock
\begin{APACrefURL} \url{https://doi.org/10.1007/s10212-011-0099-9}
  \end{APACrefURL}
\newblock
\begin{APACrefDOI} \doi{10.1007/s10212-011-0099-9} \end{APACrefDOI}
\PrintBackRefs{\CurrentBib}

\bibitem [\protect \citeauthoryear {%
Kim%
\ \BBA {} Loh%
}{%
Kim%
\ \BBA {} Loh%
}{%
{\protect \APACyear {2001}}%
}]{%
Loh01}
\APACinsertmetastar {%
Loh01}%
\begin{APACrefauthors}%
Kim, H.%
\BCBT {}\ \BBA {} Loh, W\BHBI Y.%
\end{APACrefauthors}%
\unskip\
\newblock
\APACrefYearMonthDay{2001}{}{}.
\newblock
{\BBOQ}\APACrefatitle {Classification trees with unbiased multiway splits}
  {Classification trees with unbiased multiway splits}.{\BBCQ}
\newblock
\APACjournalVolNumPages{Journal of the American Statistical
  Association}{96}{}{589--604}.
\newblock
\begin{APACrefURL}
  \url{http://www.stat.wisc.edu/~loh/treeprogs/cruise/cruise.pdf}
  \end{APACrefURL}
\PrintBackRefs{\CurrentBib}

\bibitem [\protect \citeauthoryear {%
Kononenko%
}{%
Kononenko%
}{%
{\protect \APACyear {1995}}%
}]{%
Kononenko95}
\APACinsertmetastar {%
Kononenko95}%
\begin{APACrefauthors}%
Kononenko, I.%
\end{APACrefauthors}%
\unskip\
\newblock
\APACrefYearMonthDay{1995}{}{}.
\newblock
{\BBOQ}\APACrefatitle {On Biases in Estimating Multi-valued Attributes} {On
  biases in estimating multi-valued attributes}.{\BBCQ}
\newblock
\BIn{} \APACrefbtitle {Proceedings of the 14th International Joint Conference
  on Artificial Intelligence - Volume 2} {Proceedings of the 14th international
  joint conference on artificial intelligence - volume 2}\ (\BPGS\ 1034--1040).
\newblock
\APACaddressPublisher{San Francisco, CA, USA}{Morgan Kaufmann Publishers Inc.}
\newblock
\begin{APACrefURL} \url{http://dl.acm.org/citation.cfm?id=1643031.1643034}
  \end{APACrefURL}
\PrintBackRefs{\CurrentBib}

\bibitem [\protect \citeauthoryear {%
Leeds%
\ \BBA {} DesJardins%
}{%
Leeds%
\ \BBA {} DesJardins%
}{%
{\protect \APACyear {2015}}%
}]{%
Leeds15}
\APACinsertmetastar {%
Leeds15}%
\begin{APACrefauthors}%
Leeds, D\BPBI M.%
\BCBT {}\ \BBA {} DesJardins, S\BPBI L.%
\end{APACrefauthors}%
\unskip\
\newblock
\APACrefYearMonthDay{2015}{Aug}{01}.
\newblock
{\BBOQ}\APACrefatitle {The Effect of Merit Aid on Enrollment: A Regression
  Discontinuity Analysis of Iowa's National Scholars Award} {The effect of
  merit aid on enrollment: A regression discontinuity analysis of iowa's
  national scholars award}.{\BBCQ}
\newblock
\APACjournalVolNumPages{Research in Higher Education}{56}{5}{471--495}.
\newblock
\begin{APACrefURL} \url{https://doi.org/10.1007/s11162-014-9359-2}
  \end{APACrefURL}
\newblock
\begin{APACrefDOI} \doi{10.1007/s11162-014-9359-2} \end{APACrefDOI}
\PrintBackRefs{\CurrentBib}

\bibitem [\protect \citeauthoryear {%
Liaw%
\ \BBA {} Wiener%
}{%
Liaw%
\ \BBA {} Wiener%
}{%
{\protect \APACyear {2002}}%
}]{%
Liaw02}
\APACinsertmetastar {%
Liaw02}%
\begin{APACrefauthors}%
Liaw, A.%
\BCBT {}\ \BBA {} Wiener, M.%
\end{APACrefauthors}%
\unskip\
\newblock
\APACrefYearMonthDay{2002}{}{}.
\newblock
{\BBOQ}\APACrefatitle {Classification and Regression by randomForest}
  {Classification and regression by randomforest}.{\BBCQ}
\newblock
\APACjournalVolNumPages{R News}{2}{3}{18-22}.
\newblock
\begin{APACrefURL} \url{http://CRAN.R-project.org/doc/Rnews/} \end{APACrefURL}
\PrintBackRefs{\CurrentBib}

\bibitem [\protect \citeauthoryear {%
Loh%
}{%
Loh%
}{%
{\protect \APACyear {2002}}%
}]{%
Loh02}
\APACinsertmetastar {%
Loh02}%
\begin{APACrefauthors}%
Loh, W\BHBI Y.%
\end{APACrefauthors}%
\unskip\
\newblock
\APACrefYearMonthDay{2002}{}{}.
\newblock
{\BBOQ}\APACrefatitle {Regression trees with unbiased variable selection and
  interaction detection} {Regression trees with unbiased variable selection and
  interaction detection}.{\BBCQ}
\newblock
\APACjournalVolNumPages{Statistica Sinica}{12}{}{361--386}.
\newblock
\begin{APACrefURL}
  \url{http://www.stat.wisc.edu/~loh/treeprogs/guide/guide02.pdf}
  \end{APACrefURL}
\PrintBackRefs{\CurrentBib}

\bibitem [\protect \citeauthoryear {%
Loh%
\ \BBA {} Shih%
}{%
Loh%
\ \BBA {} Shih%
}{%
{\protect \APACyear {1997}}%
}]{%
Loh97}
\APACinsertmetastar {%
Loh97}%
\begin{APACrefauthors}%
Loh, W\BHBI Y.%
\BCBT {}\ \BBA {} Shih, Y\BHBI S.%
\end{APACrefauthors}%
\unskip\
\newblock
\APACrefYearMonthDay{1997}{}{}.
\newblock
{\BBOQ}\APACrefatitle {Split selection methods for classification trees} {Split
  selection methods for classification trees}.{\BBCQ}
\newblock
\APACjournalVolNumPages{Statistica Sinica}{7}{}{815--840}.
\newblock
\begin{APACrefURL}
  \url{http://www3.stat.sinica.edu.tw/statistica/j7n4/j7n41/j7n41.htm}
  \end{APACrefURL}
\PrintBackRefs{\CurrentBib}

\bibitem [\protect \citeauthoryear {%
Mills%
\ \BBA {} Blankstein%
}{%
Mills%
\ \BBA {} Blankstein%
}{%
{\protect \APACyear {2000}}%
}]{%
Mills00}
\APACinsertmetastar {%
Mills00}%
\begin{APACrefauthors}%
Mills, J\BPBI S.%
\BCBT {}\ \BBA {} Blankstein, K\BPBI R.%
\end{APACrefauthors}%
\unskip\
\newblock
\APACrefYearMonthDay{2000}{}{}.
\newblock
{\BBOQ}\APACrefatitle {Perfectionism, intrinsic vs extrinsic motivation, and
  motivated strategies for learning: a multidimensional analysis of university
  students} {Perfectionism, intrinsic vs extrinsic motivation, and motivated
  strategies for learning: a multidimensional analysis of university
  students}.{\BBCQ}
\newblock
\APACjournalVolNumPages{Personality and Individual Differences}{29}{6}{1191 -
  1204}.
\newblock
\begin{APACrefURL}
  \url{http://www.sciencedirect.com/science/article/pii/S0191886900000039}
  \end{APACrefURL}
\newblock
\begin{APACrefDOI} \doi{http://dx.doi.org/10.1016/S0191-8869(00)00003-9}
  \end{APACrefDOI}
\PrintBackRefs{\CurrentBib}

\bibitem [\protect \citeauthoryear {%
Niessen%
, Meijer%
\BCBL {}\ \BBA {} Tendeiro%
}{%
Niessen%
\ \protect \BOthers {.}}{%
{\protect \APACyear {2016}}%
}]{%
Niessen16}
\APACinsertmetastar {%
Niessen16}%
\begin{APACrefauthors}%
Niessen, A\BPBI S\BPBI M.%
, Meijer, R\BPBI R.%
\BCBL {}\ \BBA {} Tendeiro, J\BPBI N.%
\end{APACrefauthors}%
\unskip\
\newblock
\APACrefYearMonthDay{2016}{04}{}.
\newblock
{\BBOQ}\APACrefatitle {Predicting Performance in Higher Education Using
  Proximal Predictors} {Predicting performance in higher education using
  proximal predictors}.{\BBCQ}
\newblock
\APACjournalVolNumPages{PLOS ONE}{11}{4}{1-14}.
\newblock
\begin{APACrefURL} \url{https://doi.org/10.1371/journal.pone.0153663}
  \end{APACrefURL}
\newblock
\begin{APACrefDOI} \doi{10.1371/journal.pone.0153663} \end{APACrefDOI}
\PrintBackRefs{\CurrentBib}

\bibitem [\protect \citeauthoryear {%
Ost%
}{%
Ost%
}{%
{\protect \APACyear {2010}}%
}]{%
Ost10}
\APACinsertmetastar {%
Ost10}%
\begin{APACrefauthors}%
Ost, B.%
\end{APACrefauthors}%
\unskip\
\newblock
\APACrefYearMonthDay{2010}{}{}.
\newblock
{\BBOQ}\APACrefatitle {The role of peers and grades in determining major
  persistence in sciences} {The role of peers and grades in determining major
  persistence in sciences}.{\BBCQ}
\newblock
\APACjournalVolNumPages{Economics of Education Review}{}{29}{923--934}.
\PrintBackRefs{\CurrentBib}

\bibitem [\protect \citeauthoryear {%
Sabot%
\ \BBA {} Wakeman-Linn%
}{%
Sabot%
\ \BBA {} Wakeman-Linn%
}{%
{\protect \APACyear {1991}}%
}]{%
Sabot91}
\APACinsertmetastar {%
Sabot91}%
\begin{APACrefauthors}%
Sabot, R.%
\BCBT {}\ \BBA {} Wakeman-Linn, J.%
\end{APACrefauthors}%
\unskip\
\newblock
\APACrefYearMonthDay{1991}{}{}.
\newblock
{\BBOQ}\APACrefatitle {Grade Inflation and Course Choice} {Grade inflation and
  course choice}.{\BBCQ}
\newblock
\APACjournalVolNumPages{Journal of Economic Perspectives}{5}{}{159-170}.
\PrintBackRefs{\CurrentBib}

\bibitem [\protect \citeauthoryear {%
Strobl%
, Boulesteix%
, Zeileis%
\BCBL {}\ \BBA {} Hothorn%
}{%
Strobl%
\ \protect \BOthers {.}}{%
{\protect \APACyear {2007}}%
}]{%
Strobl07}
\APACinsertmetastar {%
Strobl07}%
\begin{APACrefauthors}%
Strobl, C.%
, Boulesteix, A\BHBI L.%
, Zeileis, A.%
\BCBL {}\ \BBA {} Hothorn, T.%
\end{APACrefauthors}%
\unskip\
\newblock
\APACrefYearMonthDay{2007}{}{}.
\newblock
{\BBOQ}\APACrefatitle {Bias in random forest variable importance measures:
  Illustrations, sources and a solution} {Bias in random forest variable
  importance measures: Illustrations, sources and a solution}.{\BBCQ}
\newblock
\APACjournalVolNumPages{BMC Bioinformatics}{8}{1}{25}.
\newblock
\begin{APACrefURL} \url{http://dx.doi.org/10.1186/1471-2105-8-25}
  \end{APACrefURL}
\newblock
\begin{APACrefDOI} \doi{10.1186/1471-2105-8-25} \end{APACrefDOI}
\PrintBackRefs{\CurrentBib}

\bibitem [\protect \citeauthoryear {%
{University of Toronto}%
}{%
{University of Toronto}%
}{%
{\protect \APACyear {2017}}%
}]{%
UofT2017}
\APACinsertmetastar {%
UofT2017}%
\begin{APACrefauthors}%
{University of Toronto}.%
\end{APACrefauthors}%
\unskip\
\newblock
\APACrefYearMonthDay{2017}{}{}.
\newblock
\APACrefbtitle {Degree Requirements (H.B.A., H.B.Sc., BCom).} {Degree
  requirements (h.b.a., h.b.sc., bcom).}
\newblock
\begin{APACrefURL}
  [{2017-08-30}]\url{http://calendar.artsci.utoronto.ca/Degree_Requirements_(H.B.A.,_H.B.Sc.,_BCom).html}
  \end{APACrefURL}
\PrintBackRefs{\CurrentBib}

\end{thebibliography}

%
%

\appendix
\section{Appendix}
\label{append}

\bigskip

The following section contains some mathematical notations and definitions for readers who are interested in more a thorough explanation of sections' \ref{sectree} and \ref{secforest} content. Full understanding of the appendix is not needed in order to grasp the essential of the article but it serves as a brief but precise introduction to the mathematical formulation of decision trees and random forests. 

\bigskip

Rigorously, a typical supervised statistical learning problem is defined when the relationship between a response variable $\mathbf{Y}$ and an associated $m$-dimensional predictor vector $\mathbf{X} = (X_1,...,X_m)$ is of interest. When the response variable is categorical and takes $k$ different possible values, this problem is defined as a $k$-class classification problem. One challenge in classification problems is to use a data set $D = \{ (Y_i,X_{1,i},...,X_{m,i}) ; i = 1,...,n \}$ in order to construct a classifier $\varphi(D)$. A classifier is built to emit a class prediction for any new data point $\mathbf{X}$ that belongs in the feature space $\mathcal{X} = \mathcal{X}_1 \times ... \times \mathcal{X}_m$. Therefore a classifier divides the feature space $\mathcal{X}$ into $k$ disjoint regions such that $\cup_{j =1}^k B_l = \mathcal{X}$, i.e. $\varphi(D,\mathbf{X}) = \sum_{j=1}^k j \mathbf{1}\{ \mathbf{X} \in B_j\}$. 

\bigskip

As explained in section \ref{sectree} a classification tree \cite{Breiman84} is an algorithm that forms these regions by recursively dividing the feature space $\mathcal{X}$ until a stopping rule is applied. Most algorithms stop the partitioning process whenever every terminal node of the tree contains less than $\beta$ observations. This $\beta$ is a tuning parameter that can be established by cross-validation. Let $p_{rk}$ be the proportion of the class $k$ in the region $r$, if the region $r$ contains $n_r$ observations then : 

\begin{equation}
p_{rk}= \frac{1}{n_r} \sum_{x_i \in R_r} \mathbf{1}\{y_i = k\}.
\end{equation}

The class prediction for a new observation that shall fall in the region $r$ is the majority class in that region, i.e. if $\mathbf{X} \in R_r$, $\varphi(D,\mathbf{X}) = \textrm{argmax}_k (p_{kr})$. When splitting a region into two new regions $R_1$ and $R_2$ the algorithm will compute the total impurity of the new regions ; $ n_{1} Q_1 + n_2 Q_2$ and will pick the split variable $j$ and split location $s$ that minimizes that total impurity. If the predictor $j$ is continuous, the possible splits are of the form $X_{j} \leq s$ and $X_j > s$ which usually results in $n_r-1$ possible splits. For a categorical predictor having $q$ possible values, it is common to consider all of the $2^{q-1} -1$ possible splits. Hastie \& al. \citeyear{Hastie09} introduces many possible region impurity measurements $Q_r$, in this project, the \textit{Gini index} has been chosen :

\begin{equation}
Q_r = \sum_{j=1}^k p_{rj}(1-p_{rj}).
\end{equation}

Here is a pseudo-code of the algorithm :
 
\begin{center}
\begin{tabular}{||l||}
\hline
\textbf{Algorithm} : DT($D$,$\beta$) \\
\hline
\hline
1. Starting with the entire data set $D$ as the first set of observations $r$.  \\
2. Check ($n_r$ > $\beta$). \\
3. \textbf{if}  (false) :  \\
\hspace{0.5cm} Assign a label to the node and exit.  \\
\textbf{ else if}  :  \\
\hspace{0.5cm} \textbf{for} ($j$ in all predictors):  \\
\hspace{1.0cm} \textbf{for} ($s$ in all possible splits) : \\
\hspace{1.5cm} Compute total impurity measure. \\
\hspace{0.5cm} Select variable $j$ and split $s$ with minimum impurity measure and split \\
\hspace{0.5cm} the set $r$ into two children sets of observations.\\
\hspace{0.5cm} Repeat steps 2 \& 3 on the two resulting sets.\\ 
\hline
\end{tabular}
\end{center}

\bigskip

Since decision trees are unstable procedures \cite{Breiman96a} they greatly benefit from bootstrap aggregating (bagging) \cite{Breiman96}. In classifier aggregating, the goal is to find a way to use an entire set of classifiers $\{ \varphi(D_q) \}$  to get a new classifier $\varphi_a$ that is better than any of them individually.  One method of aggregating the class predictions $\{ \varphi(D_q,\mathbf{X}) \}$ is by \textit{voting}: the predicted class for the input $\mathbf{X}$ is the most picked class among the classifiers. More precisely, let $T_k = | \{ q : \varphi(D_q, \mathbf{X}) = k \} |$ then, the aggregating classifier becomes $\varphi_a(\mathbf{X}) = \textrm{argmax}_k (T_k)$. 

\bigskip

On way to form a set of classifiers is to draw bootstrap samples of the data set $D$ which forms a set of learning sets $\{ D_B \}$. Each of the bootstrap samples will be of size $n$ drawn at random with replacement from the original training set $D$. For each of these learning set a classifier  $\varphi(D_b)$ is constructed and the resulting set of classifiers $\{ \varphi(D_b) \}$ can be used to create an aggregating classifier. If the classifier is an unpruned tree then the aggregating classifier is a random forest.

\bigskip
 
A random forest classifier is more precise than a single classification tree in the sense that it has lower mean-squared prediction error \cite{Breiman96}. By bagging a classifier, the bias will remain the same but the variance will decrease. One way to further decrease the variance of the random forest is by construction trees that are as uncorrelated as possible. Breiman introduced in 2001 random forests with random inputs \cite{Breiman01}. In these forests, instead of finding the best variable and partitioning among all the variables, the algorithm will now randomly select $p < m$ random covariates and will find the best condition among those $p$ covariates. 

\bigskip

The fitted random forest classifiers were compared to two logistic regression models. A simple logistic model is used to predict if a student completes its program or not with the following parametrization :

\begin{equation}
P(Y_i =1)   = \frac{\exp(\sum_{i=0}^m \beta_i x_i)}{1+\exp(\sum_{i=0}^m \beta_i x_i)},
\end{equation}
where $Y_i=1$ means student $i$ completed its program, $m$ is the number of predictors, $\beta's$ the parameters and $x_i's$ the predictor values. To predict the major completed, a generalization of the logistic regression, the multinomial logistic regression is used with the following parametrization :

\begin{equation}
P(Y_i = p)   = \frac{\exp(\sum_{i=0}^m \beta_i^{(p)} x_i)}{1+\exp(\sum_{l=1}^k \sum_{i=0}^m \beta_i^{l} x_i)},
\end{equation}
where $Y_i =p$ means the student $i$ completed the program $p$ and where $k$ is the number of programs. 

\bigskip

Finally, here is a short example of code to fit random forests, get predictions for new observations and produce variable importance plots using the R language :

\begin{verbatim}
#Importing the randomForest package
require(randomForest)

#Fitting the random forest with 200 trees
#using bootstraps without replacement.
Fit <- randomForest(x=X,y=as.factor(Y),importance=TRUE,ntree=200,
replace=FALSE,sampsize=round(0.63*nrow(X)) )

#Prediction class labels for new observations newX
predictions <- predict(Fit,newX)

#Production variable importance plot
importance(Fit,type=1)

\end{verbatim}

\end{document}